\documentclass[runningheads]{llncs}
\usepackage{graphicx}
\usepackage{comment}
\usepackage{enumitem}
\usepackage{algorithm}
\usepackage{multirow}
\usepackage{caption}
\usepackage{listings}
\usepackage{lscape}
\usepackage{rotating}
\usepackage{seqsplit}
\usepackage[noend]{algpseudocode}
\usepackage[fleqn]{amsmath}

\newcommand\blfootnote[1]{%
  \begingroup
  \renewcommand\thefootnote{}\footnote{#1}%
  \addtocounter{footnote}{-1}%
  \endgroup
}
   
\begin{document}

\title{ChatGPT as a commenter to the news:\\ can LLMs generate human-like opinions?
}

\titlerunning{ChatGPT as a commenter to the news}
%
\author{Rayden Tseng\inst{1} \and Suzan Verberne\inst{1}\orcidID{0000-0002-9609-9505} \and Peter van der Putten\inst{1}\orcidID{0000-0002-6507-6896}
}
\authorrunning{Tseng et al.}
%

\institute{
LIACS, Leiden University, Leiden, the Netherlands\\\email{{p.w.h.van.der.putten,s.verberne}@liacs.leidenuniv.nl}\\}

\maketitle              
\begin{abstract}
ChatGPT, GPT-3.5, and other large language models (LLMs) have drawn significant attention since their release, and the abilities of these models have been investigated for a wide variety of tasks. In this research we investigate to what extent GPT-3.5 can generate human-like comments on Dutch news articles. We define human likeness as `not distinguishable from human comments', approximated by the difficulty of automatic classification between human and GPT comments. We analyze human likeness across multiple prompting techniques. In particular, we utilize zero-shot, few-shot and context prompts, for two generated personas. We found that our fine-tuned BERT models can easily distinguish human-written comments from GPT-3.5 generated comments, with none of the used prompting methods performing noticeably better. We further analyzed that human comments consistently showed higher lexical diversity than GPT-generated comments. This indicates that although generative LLMs can generate fluent text, their capability to create human-like opinionated comments is still limited.

\end{abstract}

\section{Introduction}

\blfootnote{Paper presented as Tseng, R., Verberne, S., van der Putten, P. (2023). ChatGPT as a Commenter to the News: Can LLMs Generate Human-Like Opinions?. In: Ceolin, D., Caselli, T., Tulin, M. (eds) Disinformation in Open Online Media. MISDOOM 2023. Lecture Notes in Computer Science, vol 14397. Springer, Cham. 
}

Since the public availability of GPT-3.5, its capabilities have been researched for a wide range of tasks \cite{brown2020language}. It has shown remarkable performance in text summarization \cite{zhang2023extractive}, machine translation \cite{hendy2023good} and classification, such as hate speech detection \cite{chiu2021detecting} and sentiment analysis \cite{wang2023chatgpt}. GPT-2 has previously been used for generating fake product reviews  \cite{Adelani2020,Salminen2022}, and GPT-3.5 to generate Tweets 
\cite{spitale2023ai}. To our knowledge, there is no research into to what extent GPT-3.5 can generate human-like opinions on news articles. 

In this paper, we present a small-scale study investigating the capability of GPT-3.5 to produce opinionated text, and more specifically, comments on news articles. For the purpose of this study, we loosely define `human-like' as `not distinguishable or difficult to distinguish from human'. We crawled human comments from a Dutch newspaper website, and generated opinions by prompting GPT with two different generated personas. We supplied just the title or additional context from the article, and in zero-shot and few-shot settings. Subsequently, we analyze to what extent the GPT-3.5 generated comments can be distinguished from human comments, using a fine-tuned Dutch BERT model.
As additional analysis, we use other metrics to investigate the differences between the outputs, such as type-token ratio and qualitative analysis of misclassifications with SHAP. 

\section{Related work}\label{sec:relwork}

\subsection{Large Language Models and GPT-3.5}
Large Language Models are Transformer-based language models, which enables them to capture contextual dependencies and generate human-like text. It was shown that when models with a large amount of parameters are trained on large amounts of text, they can solve problems for a wide range of tasks. 
The Generative Pre-trained Transformer 3.5 (GPT-3.5) from OpenAI 
has 175 billion parameters
, and is trained on approximately 499 billion tokens \cite{brown2020language}. Its companion GPT-4 model is even more powerful, even though architectural details have not been made public. The architecture of GPT is decoder-only, which allows for open-ended generation. Unlike encoder-decoder architectures, the output of decoder-only models is less scoped by the input, which enlarges the space of acceptable output generations and therefore increases its potential for more diverse and creative responses \cite{holtzman2019curious}. Without any fine-tuning or gradient updates, it has shown strong performance in many NLP tasks on many datasets.

\subsection{BERT}
In 2019, Devlin et al. introduced a Transformer-based language model called BERT, which stands for Bidirectional Encoder Representations from Transformers \cite{devlin2019bert}. In contrast to GPT, the BERT model is a transformer \emph{encoder}, without a decoder component. 
BERT's pre-training makes use of a Masked Language Modeling (MLM) objective. It randomly masks some tokens from the input, after which its objective is to predict the original token based on the context. 
in addition, it uses Next Sentence Prediction (NSP), which, given two sentences, determines whether the second sentence follows the first. MLM and NSP help BERT understand context across different sentences. 
BERT models can be fine-tuned for specific tasks, creating new state-of-the-art models for supervised learning tasks. 

RobBERT is a large pre-trained general Dutch language model that can be fine-tuned on a given dataset to perform a wide range of NLP tasks \cite{delobelle2020robbert}. It uses a RoBERTa architecture and pre-training with a Dutch tokenizer. RoBERTa, a Robustly Optimized BERT Pre-training Approach has some modified key hyperparameters, such as removing the NSP objective in pre-training, having much larger batches and training on longer sequences \cite{liu2019roberta}. It has shown state-of-the-art results for various tasks, especially compared to other models when applied to smaller datasets \cite{delobelle2020robbert}. In this research, we use the Dutch RobBERT model to fine-tune for our specific use case, classifying comments as GPT- or human-generated.

\subsection{Previous research on GPT-3.5's capabilities}
Since the public release of GPT-3.5 its capabilities have been researched extensively.
For summarization, it was shown that its extractive summarization performance is inferior compared to existing methods  \cite{zhang2023extractive}, but other research showed that GPT models have competitive translation performance for high-resource languages, while still being limited for low-resource languages 
\cite{hendy2023good}. In addition to its capabilities to produce text, GPT has also been evaluated for its ability to classify. Chiu et al. used zero-shot and few-shot prompting techniques to investigate whether GPT-3 can identify sexist or racist text. They found an average zero-shot accuracy between 55\% and 67\%, whereas few-shot learning achieved an accuracy that can be as high as 85\% \cite{chiu2021detecting}. 
Another study 
has shown that ChatGPT exhibits impressive zero-shot performance in sentiment classification tasks and can rival a fine-tuned BERT model. Few-shot learning further enhances its performance, even surpassing fine-tuned BERT models in some cases \cite{wang2023chatgpt}.

In terms of language generation in online media, GPT-3 has also been evaluated on whether it could write human-like content on social media through the form of tweets. In this research, human participants were asked to determine whether tweets were human-written or machine-generated. This has shown that GPT-3, in comparison with humans, can produce accurate (dis)information that is easier to understand and that humans could not distinguish whether tweets were generated or written by humans \cite{spitale2023ai}. A task even closer to our use case is the generation of fake product reviews, and it was demonstrated that GPT-2 can already be used to generate such reviews, but also that in some cases classifiers could be built to detect these reviews more successfully than humans \cite{Adelani2020,Salminen2022}.

\section{Methods}\label{sec:methods}
In this paper, we address the capabilities of GPT-3.5 for producing opinionated comments on news articles. We evaluate this by fine-tuning a BERT model on the task of classifying comments as either Human- or GPT-generated. In this section, we describe the methods that were used for this research. 
This includes (1) collecting news articles and human comments from an online platform; (2) generating opinions with GPT-3.5 using two generated personas, zero-shot and few-shot settings, and using just the article title or also the introduction as context in the prompt; (3) assessing the quality of the output by evaluating how well a fine-tuned Dutch-based BERT model is able to detect generated opinions. All code is written in Python.

\subsection{Data collection}
In order to compare human to artificial opinions, we collect opinions on news articles from \emph{NU.nl}, a major online news channel in the Netherlands, with a politically centrist and reporting oriented positioning. It features an integrated comment system \emph{NUjij}.
We collected a total of ten articles, each containing at least a hundred comments. The articles and their corresponding numbers of comments (opinions) are shown in Table \ref{tab:articles}. 
The main requirement of an appropriate article was that it discussed a topic on which opinions generally differ. We selected articles with at least 100 comments. Since an account was required to access the comments, each HTML page needed to be downloaded manually. For this research, the Chrome extension Save Page WE was used after having registered to the platform.\footnote{\url{https://chrome.google.com/webstore/detail/save-page-we/dhhpefjklgkmgeafimnjhojgjamoafof}} 
 
\begin{table}[t]
\centering
\resizebox{\columnwidth}{!}{%
\begin{tabular}{c|l|c|l}
\textbf{Number} & \multicolumn{1}{c|}{\textbf{Title}}                                              & \multicolumn{1}{l|}{\textbf{Opinions}} & \multicolumn{1}{c}{\textbf{Date}} \\ \hline
\textbf{1}   & Avondklok besproken als `serieuze optie', maar invoering nog niet aan de orde      & 103                                    & 12/01/2021                        \\
\textbf{2}   & Burgemeester Parijs: `Geen Russische atleten op Spelen zolang oorlog woedt'        & 143                                    & 03/02/2023                        \\
\textbf{3}   & EU adviseert QR-code tot 9 maanden na laatste prik te laten gelden voor reizen     & 101                                    & 25/11/2021                        \\
\textbf{4}   & Feyenoord-aanvoerder Kökçü weigert vanwege religie regenboogband te dragen         & 194                                    & 16/10/2022                        \\
\textbf{5}   & Jumbo stopt per direct met WK-reclamespot na storm van kritiek                     & 144                                    & 02/11/2022                        \\
\textbf{6}   & Minister rekent op 1.400 euro vergoeding voor studenten uit `pechgeneratie'        & 104                                    & 25/03/2022                        \\
\textbf{7}   & Rusland valt Oekraïne aan, oorlog breekt uit                                       & 105                                    & 24/02/2022                        \\
\textbf{8}   & Rutte biedt excuses aan voor slavernijverleden:  `Aan alle nazaten tot hier en nu' & 160                                    & 19/12/2022                        \\
\textbf{9}   & Studentenorganisaties willen tijdelijke rem op komst internationale studenten      & 122                                    & 02/02/2023                        \\
\textbf{10}  & Talpa wist volgens BOOS mogelijk al veel langer van misstanden bij The Voice       & 110                                    & 25/08/2022   \\                    
\hline
\end{tabular}%
}
\caption{Titles of articles with number of comments and publication dates}
\label{tab:articles}
\end{table}

\textbf{Human opinions}
Once the articles had been downloaded and stored in the same directory, we parsed the HTML content to extract the article text and comments. Our implementation only included parent comments since sub-comments might deviate from the initial topic. We found that the comments were contained in the \emph{coral-comment-content} class and the text in the \emph{textblock paragraph} class. Firstly, using the BeautifulSoup\footnote{\url{https://www.crummy.com/software/BeautifulSoup/bs4/doc/}} library, the text was extracted and secondly the comments. Subsequently, both the text and comments were individually written to files in newly created directories. It appeared that the comments inside the HTML pages had an inconsistent structure, resulting in varying outcomes during scraping. While some articles were scraped without any issues, others showed HTML tags or unusual punctuation. This includes unusual \verb|<br>|, \verb|<br/>|, \verb|<div>| or \verb|<p>| tags for no apparent reason, alongside inconsistent use of \verb|'| or \verb|"|. Since there was no other way to resolve this, all inconsistencies had to be corrected manually. 
All ten articles had between 100 and 200 comments. Since GPT-3.5's pre-training data is from before September 2021, an important remark here is that all articles except for article 1 were published after this date. 

\subsection{Generating opinions with GPT-3.5}
\label{subsec:method}
After correctly parsing the text and opinions of the articles, opinions were generated. Using the \emph{gpt-3.5-turbo} model, we were able to take advantage of a longer input context and incorporate human responses into the conversation as examples, whereas earlier models like \emph{text-davinci-003} could not. Since each article contains at least 100 human opinions, the goal was to generate 100 artificial opinions per article. 

\textbf{Prompts and personas}
First of all, prompts had to be constructed. By providing the title of the article in the prompt, we first ran a set of test prompts to explore the quality of the responses of the model. It became evident that the model 
mostly gave formal, boring and factual perspectives on the subject. 

To generate more opinionated and human-like content, we used GPT-3.5 to generate personas, such that subsequently we could generate opinions through a specific perspective, expecting more personalization in the opinions. 
Two random personas were generated in advance and provided in this new contextual model. Obviously, two is a limited number of personas, but we wanted to explore the potential of persona-based opinion generation, and also see whether both the output as well as the detection would vary across personas.

This was accomplished by appending the personas to the \verb|system role|, such that the model knew how to `behave'. Every time the experiments were run, we prompted GPT-3.5 to generate these random personas. We ultimately came up with the following prompt: \texttt{Generate a persona. Use three sentences. Start with `You are'.} This prompt was constructed based on several criteria. Firstly, it was necessary to be concise and easily be easily understood. Secondly, since it would be used in other prompts to experiment with other settings, the output had to start with \texttt{`You are'}. Other options such as providing more information about what kind of persona it had to generate were also taken into consideration. We tried to provide the information that this persona likes to read an online news platform. However, after evaluating this, it became evident that it resulted in a more general persona based on the fact that it likes to read, rather than characteristics which would influence its opinion. It was nevertheless important to provide this information. We appended \texttt{`You comment on an online newsplatform'} to the output manually.
At last, we could pass this information to the other settings by appending the persona to the \texttt{role} contents. 
In Section~\ref{sec:results}, these personas are referred to as 1 and 2 or P1 and P2.

Another observation after analyzing the test completions was that it consisted of opinions that were significantly shorter in length than human opinions. To address this, we provided an approximate length for each opinion in the prompt, calculated for each individual article. We used the average length of human comments on the article for determining the comment length. In total, we investigated four settings, described below.

\textbf{Zero-shot}
The first setting was designed to utilize zero-shot learning. Recent work has shown that large language models exhibit the ability to perform reasonable zero-shot generalization to new tasks \cite{sanh2021multitask}. In this case of generating human-like perspectives, the model was solely prompted to generate opinions, with the title of the article as additional information. With this approach, we could examine GPT's creativity to the fullest. However, depending on the topic of the article, the model generally has the least amount of context with this prompt. This led to the following prompt:
  \begin{verbatim}
    Give a list of 100 varied and critical opinions on the 
    following news article: `w', where each opinion has an 
    approximate length of `x' words.\end{verbatim}
Here, \verb|w| refers to the title of article and \verb|x| to the approximate length of the comment. In Section~\ref{sec:results}, the zero-shot prompt is referred to as ZS.

\textbf{Few-shot}
In the second setting, a few examples of human opinions were provided. The aim was to give the model more context and guidance to generate opinions similarly. Due to the limitations discussed in Section~\ref{sec:limit1}, only four examples are provided. Assuming the relevance is high, the comments with the most likes were selected. This method was most likely to perform best. It first of all learned directly from real examples and therefore might adapt its style and tone more accurately. This setting resulted in the following prompt:

\begin{verbatim}
    Give a list of 100 varied and critical opinions on the 
    following news article: `w', where each opinion has an 
    approximate length of `x' words. Here are four examples: `y'.
    \end{verbatim}
\vspace{-0.5cm} 
In the prompting script, the variable \verb|y| was replaced by the first four examples of the article. Later on, the few-shot approach is described as FS.

\textbf{Context}
In the third setting the experiments were run with is additional context. In addition to the zero-shot prompt, the introduction of the article was provided. The model, therefore, had more context to work with. This method may be beneficial since GPT-3.5 can produce more in-depth opinions, whereas the zero-shot prompt or few-shot can not. A potential downfall is that the introduction might not always contain any relevant or additional information. This resulted in the final prompt:

\begin{verbatim}
    Give a list of 100 varied and crital opinions on the following 
    news article: `w', where each opinion has an approximate length 
    of `x' words. This is the introduction of the article: `z'.
    \end{verbatim}
\vspace{-0.5cm}  The \verb|z| variable was substituted by the introduction of the corresponding article. The context prompt is later described as CL. We utilized ten different articles and three different prompting techniques, each with two different personas. Therefore, the number of total prompts is equal to 60 ($10 \times 3 \times 2$). 

\textbf{Limitations}
\label{sec:limit1}
In practice however, there were some limitations. Language models read text in chunks called tokens. Tokenization is the process of splitting text into smaller units called tokens, which is a fundamental preprocessing step for almost all NLP tasks \cite{song2020fast}. Due to a maximum request of 4,096 tokens, which includes both the prompt and completion, the model would stop generating when that limit was reached. This, in particular for the few-shot and context prompt, resulted in unfinished completions. This limitation was also the reason we could not provide more examples or even the entire article in the prompt, which in theory would perform better. We came up with a solution by reducing the request of opinions to 25 (and 20 for the few-shot setting), such that the request was more likely to stay within the limit and increasing the number of requests per prompt to 4 (5 for few-shot).

\subsection{Evaluation through classification}
After we successfully prompted GPT-3.5, with 6 modes and ten articles we ideally would have 6000 generated opinions. To best evaluate the difference between the output of the prompts, different classification models were built. The data first needed to be correctly formatted such that the BERT model could process it. A small Python script was written, which iterates over all the outputs and then formats them to a readable database format. For this implementation, the .csv file format was chosen. An important remark here is that each record consists of a text \emph{comment} and a \emph{human} column, indicating the boolean value for Human- (1) or AI-generated (0).
The data must be split into train and test sets. We split the data on article level, keeping all comments to the same article together in one partition. We processed the data in such a way, that we could examine an individual setting with every possible article as a test set. This in total, resulted in 120 files, where each file was either a train or test set, given a setting and article.
Due to the lack of computational power available, the classification phase is performed using Google Colab.

\textbf{Fine-tuning}
We analyze the human likeness of the generated comments by classifying them. 
We fine-tuned \textit{robbert-v2-dutch-base}, which is the state-of-the-art Dutch BERT model \cite{delobelle2020robbert}. The goal of the models was to predict whether a given opinion is human-generated or GPT-3.5-generated.
To ensure the models were properly trained, as mentioned earlier, we fine-tuned them via 10-fold cross-validation. Here for each fold, we trained the model on the other 9 of the 10 article sets and evaluated its performance on the remaining set. In this research, 10 models were trained for each setting, rather than a single model. We made this choice to prevent the model from overfitting the training data and to ensure it could generalize well to unseen data. 

To fine-tune the BERT model, several steps had to be taken. First, the generated train and test sets needed to be imported. In order for the model to read the input, the input text needed to be tokenized using the tokenizer from the pre-trained model. Since we did not use an existing dataset from the HuggingFace library, our tokenized data had to be converted to a suitable \textit{Dataset} object, in order for the trainer to run without errors.

Before training the model, the \emph{TrainingArguments} needed to be initialized. TrainingArguments are a subset of arguments that relate to the training process. For this analysis, we used the default values provided by the HuggingFace tutorial.\footnote{\url{https://huggingface.co/docs/transformers/training}} Specifically, the learning rate at which the model adapts its parameters while training was set to $2\times10^{-5}$ and the batch size per training core to 8. The weight decay was set to 0.01, while the evaluation and save strategies were configured to epoch. The number of training epochs was set to 3.

To evaluate the model, we used the F1 score, recall and precision as metrics. These methods can be imported via the Python \emph{evaluate} library. In addition to these metrics, we also created a confusion matrix.

\section{Results} \label{sec:results}

The results of the research are discussed in this section. It contains an overview of the output of the prompts and the metrics of the fine-tuned models. Also, we analyze the lexical diversity of the output discuss two examples qualitatively. 

\subsection{Output}
Before utilizing all prompts, the two personas were generated, which resulted in the following two completions (translated to English):

\texttt{
You are a middle-aged man, married and father of two children. You 
have worked as an accountant at a large company for 20 years and you
enjoy playing tennis in your spare time. You comment on an online 
news platform.}

and

\texttt{
You are a 32-year-old marketing executive who loves socialising and travelling. You have a busy job, but find it important to spend time with family and friends. In your spare time, you like to be outdoors and do running and yoga. You have a passion for cooking and are always trying out new recipes. You comment on an online news platform.}

Ideally, the methods described in Section~\ref{subsec:method} would result in 100 comments per setting per article. In practice, a total of 5855 generated comments were generated, where a few prompts did not result in the full completion due to the limitations. As a consequence, balanced datasets could not be generated in some cases. We wanted to achieve this anyway, so in some cases, we had to generate the remaining responses manually, by prompting GPT-3.5 separately from the script. In Table~\ref{tab:generated_comments}, the article, setting and number of generated comments are shown. We also released all generated comments on github.\footnote{\url{https://github.com/raydentseng/generated\_opinions}}

\begin{table}[t]
\centering
\begin{tabular}{c|cccccc|c}
\hline
\textbf{Article} & \textbf{ZS-1} & \textbf{FS-1} & \textbf{CL-1} & \textbf{ZS-2} & \textbf{FS-2} & \textbf{CL-2} & \textbf{Total} \\ \hline
\textbf{1}       & 100           & 92            & 100           & 100           & 73            & 100           & 565            \\
\textbf{2}       & 100           & 95            & 100           & 100           & 100           & 100           & 595            \\
\textbf{3}       & 100           & 71            & 100           & 100           & 75            & 100           & 546            \\
\textbf{4}       & 100           & 73            & 100           & 100           & 100           & 100           & 573            \\
\textbf{5}       & 100           & 88            & 100           & 100           & 100           & 100           & 588            \\
\textbf{6}       & 100           & 100           & 100           & 100           & 100           & 100           & 600            \\
\textbf{7}       & 100           & 100           & 100           & 100           & 88            & 100           & 588            \\
\textbf{8}       & 100           & 100           & 100           & 100           & 100           & 100           & 600            \\
\textbf{9}       & 100           & 100           & 100           & 100           & 100           & 100           & 600            \\
\textbf{10}      & 100           & 100           & 100           & 100           & 100           & 100           & 600            \\ \hline
\textbf{Total}   & 1000          & 919           & 1000          & 1000          & 936           & 1000          & 5855          \\
\hline
\end{tabular}
\caption{Amount of generated comments per setting. A setting consists of a prompt and a generated persona. ZS refers to the zero-shot prompt, FS to few-shot and CL to the context prompt. For instance, FS-2 indicates few-shot with persona 2.}
\label{tab:generated_comments}
\end{table}

On preliminary manual analysis of the completions, a few things immediately stood out. Firstly, it became evident that each prompt did not complete in a single run. Each prompt, despite being the same, consisted of multiple requests, resulting in multiple batches. This was evident as there was less variation within one batch than between different batches. Batches showed clear differences in overall writing style. It appears that the output relies on earlier produced tokens in the same request. 
A second preliminary observation we made was the nature of the generated comments. While these could be classified as opinions, they exhibited mostly factual text or reasoned arguments rather than expressing emotional viewpoints, even having utilized the few-shot prompt. Besides, the comments looked rather formal, as opposed to the human-written comments which often contained more informal language and slang.


\subsection{Classification results}

Our main results are presented in Table~\ref{tab:mainresults}. Each row contains a single setting. The columns represent the average and standard deviation of metrics of all the fine-tuned models. All values in Table~\ref{tab:mainresults} are relatively high. Initially, this seems positive. However, we are most interested in the lowest scores, since it reveals the cases where the model encountered the most challenges in distinguishing human-written comments from machine-generated comments. Lower values therefore indicate better performance of that setting. In terms of the F1-score, which represents the overall performance of the model, there is little difference, with the ZS-1 setting having the lowest of 91.2\%. However, it does not have the lowest precision and recall of both classes, and the differences to other settings are small. 

\begin{table}[t]
\centering
\resizebox{\columnwidth}{!}{%
\begin{tabular}{c|c|cc|cc}
\hline
\textbf{}                             & \textbf{}              & \multicolumn{2}{c}{\textbf{GPT-3.5}}            & \multicolumn{2}{|c}{\textbf{Human}}          \\ 
\multicolumn{1}{c|}{\textbf{Setting}} & \textbf{F1}            & \textbf{Precision}     & \textbf{Recall}        & \textbf{Precision} & \textbf{Recall}        \\ \hline
\multicolumn{1}{c|}{\textbf{ZS-1}}    & \textbf{0.912 ± 0.047} & 0.926 ± 0.055          & 0.924 ± 0.084          & 0.931 ± 0.065      & 0.912 ± 0.065          \\
\multicolumn{1}{c|}{\textbf{FS-1}}    & 0.936 ± 0.032          & 0.943 ± 0.039          & 0.914 ± 0.082          & 0.939 ± 0.065      & 0.940 ± 0.044          \\
\multicolumn{1}{c|}{\textbf{CL-1}}    & 0.925 ± 0.043          & 0.932 ± 0.046          & 0.922 ± 0.103          & 0.932 ± 0.080      & 0.928 ± 0.056          \\
\multicolumn{1}{c|}{\textbf{ZS-2}}    & 0.923 ± 0.024          & 0.937 ± 0.041          & 0.918 ± 0.039          & \textbf{0.913 ± 0.038}      & 0.936 ± 0.045          \\
\multicolumn{1}{c|}{\textbf{FS-2}}    & 0.934 ± 0.027          & 0.962 ± 0.027          & \textbf{0.896 ± 0.074} & 0.915 ± 0.056      & 0.962 ± 0.030          \\
\multicolumn{1}{c|}{\textbf{CL-2}}    & 0.920 ± 0.034          & \textbf{0.888 ± 0.047} & 0.966 ± 0.071          & 0.969 ± 0.062      & \textbf{0.873 ± 0.090} \\
\hline
\end{tabular}%
}
\caption{Results per setting for both classes Human and GPT-3.5. Lower scores mean that classification was more difficult, meaning that the GPT-generated comments were more human-like.}
\label{tab:mainresults}
\end{table}

In Table~\ref{tab:metricsarticle}, the metrics are shown per article. Just like in Table~\ref{tab:mainresults}, all values are close to each other. The F1 scores are again all around 90\%. Overall, models which used article 1 as test set resulted in the lowest F1 score of 89.2\%. This is interesting because article 1 is the only article that was published before the pre-training date of GPT-3.5. In other words, the article is the only topic (Covid regulations in the Netherlands) that GPT-3.5 likely has covered in its pre-training data.

\begin{table}[t]
\centering
\resizebox{\columnwidth}{!}{%
\begin{tabular}{c|c|cc|cc}
\hline
\textbf{}                             & \textbf{}              & \multicolumn{2}{c}{\textbf{GPT-3.5}}            & \multicolumn{2}{|c}{\textbf{Human}}              \\ 
\multicolumn{1}{c|}{\textbf{Article}} & \textbf{F1}            & \textbf{Precision}     & \textbf{Recall}        & \textbf{Precision}     & \textbf{Recall}        \\ \hline
\multicolumn{1}{c|}{\textbf{1}}       & \textbf{0.892 ± 0.029} & 0.971 ± 0.008          & \textbf{0.812 ± 0.085} & \textbf{0.843 ± 0.061} & 0.975 ± 0.008          \\
\multicolumn{1}{c|}{\textbf{2}}       & 0.928 ± 0.008          & 0.906 ± 0.028          & 0.958 ± 0.025          & 0.957 ± 0.024          & 0.898 ± 0.036          \\
\multicolumn{1}{c|}{\textbf{3}}       & 0.931 ± 0.023          & 0.947 ± 0.019          & 0.913 ± 0.050          & 0.919 ± 0.044          & 0.948 ± 0.019          \\
\multicolumn{1}{c|}{\textbf{4}}       & 0.945 ± 0.017          & 0.930 ± 0.031          & 0.965 ± 0.008          & 0.963 ± 0.010          & 0.925 ± 0.037          \\
\multicolumn{1}{c|}{\textbf{5}}       & 0.913 ± 0.028          & 0.954 ± 0.019          & 0.870 ± 0.083          & 0.887 ± 0.066          & 0.957 ± 0.019          \\
\multicolumn{1}{c|}{\textbf{6}}       & 0.937 ± 0.021          & 0.979 ± 0.016          & 0.893 ± 0.067          & 0.907 ± 0.055          & 0.980 ± 0.017          \\
\multicolumn{1}{c|}{\textbf{7}}       & 0.926 ± 0.014          & 0.886 ± 0.026          & 0.982 ± 0.024          & 0.981 ± 0.024          & 0.872 ± 0.034          \\
\multicolumn{1}{c|}{\textbf{8}}       & 0.927 ± 0.019          & 0.910 ± 0.037          & 0.952 ± 0.028          & 0.951 ± 0.028          & 0.903 ± 0.043          \\
\multicolumn{1}{c|}{\textbf{9}}       & 0.914 ± 0.010          & \textbf{0.867 ± 0.028} & 0.980 ± 0.017          & 0.978 ± 0.018          & \textbf{0.848 ± 0.038} \\
\multicolumn{1}{c|}{\textbf{10}}      & 0.948 ± 0.013          & 0.961 ± 0.035          & 0.938 ± 0.045          & 0.943 ± 0.0395         & 0.958 ± 0.039     \\
\hline
\end{tabular}%
}
\caption{Metrics per article. Lower scores mean that classification was more difficult, meaning that the GPT-generated comments were more human-like.}
\label{tab:metricsarticle}
\end{table}

\subsection{Lexical Diversity}

Another quantitative method to analyze the different outputs is the \emph{Type-Token Ratio} (TTR). TTR is the ratio calculated by dividing the types (\emph{t}), which are the unique words occurring in a text, by its tokens (\emph{n}), the total number of words. This, therefore, measures the lexical diversity, given multiple texts. 
After some preliminary observations of the generated output, we noticed that it seemed that GPT-3.5 used a lot of the same words. By utilizing the LexicalRichness\footnote{\url{https://github.com/lsys/lexicalrichness}} Python package~\cite{lexicalrichness}, the lexical variety between human comments and GPT-3.5 generated comments can be quantitatively measured. Since not all generated comments equalled as many tokens as human comments and longer texts tend to have higher TTR values because they have more opportunities for unique words to occur, we used the \emph{Corrected Type-Token Ratio} (CTTR) \cite{torruella2013lexical} metric. CTTR normalizes the TTR by using the square root, providing a more accurate measure by considering the potential effect of the text length, and is calculated as $\frac{t}{\sqrt{2n}}$, where $t$ is the number of unique terms in a text and $n$ is the total number of tokens. We computed the CTTR value over the total text of concatenated comments. In Table~\ref{tab:cttr}, the CTTR of all human and generated comments are shown. Since the number of human comments differs significantly per article, the first 100 are taken into account. The highest calculated GPT-3.5 CTTR per article is boldfaced. It appears the human text consistently has a higher value, compared to all the different settings. None of the settings matches the value of the human CTTR. From all generated completions, the few-shot completion of article 5 had the highest calculated CTTR value.  

\begin{table}[t]
\centering
\begin{tabular}{c|c|cccccc}
\hline
\textbf{Article} & \textbf{Human} & \textbf{ZS-1}  & \textbf{FS-1}  & \textbf{CL-2}  & \textbf{ZS-2}  & \textbf{FS-2}   & \textbf{CL-2} \\ \hline
\textbf{1}       & 14.422         & 8.892          & 8.707          & 9.104          & 8.664          & \textbf{9.794}  & 9.218         \\
\textbf{2}       & 13.043         & \textbf{9.103} & 8.559          & 7.531          & 7.569          & 8.650           & 7.940         \\
\textbf{3}       & 13.192         & 9.062          & 9.893          & \textbf{9.833} & 9.442          & 8.973           & 9.196         \\
\textbf{4}       & 10.581         & 7.778          & 8.697          & \textbf{8.970} & 8.493          & 7.822           & 8.198         \\
\textbf{5}       & 12.457         & 9.018          & 9.503          & 8.302          & 8.302          & \textbf{10.191} & 9.641         \\
\textbf{6}       & 13.657         & 8.126          & 9.334          & 7.458          & 8.671          & \textbf{10.027} & 7.751         \\
\textbf{7}       & 14.436         & 9.043          & \textbf{9.056} & 8.620          & 8.370          & 8.247           & 7.837         \\
\textbf{8}       & 12.963         & 7.831          & 7.878          & 7.768          & 7.221          & \textbf{8.674}           & 6.738         \\
\textbf{9}       & 13.552         & 9.296          & 8.643          & 7.341          & \textbf{9.334} & 8.133           & 6.862         \\
\textbf{10}      & 13.440         & 6.330          & 7.639          & 7.318          & 7.410          & \textbf{8.093}  & 6.819     \\
\hline
\end{tabular}
\caption{CTTR values per article. Boldface indicates the value for the GPT-3.5 model that is closest to the Human value for the article.}
\label{tab:cttr}
\end{table}

\subsection{Qualitative analysis}
Aside from the quantitative analysis, we can also analyze the output qualitatively. Since our fine-tuned model classified 0 as \emph{AI} (\emph{GPT-3.5}) and 1 as \emph{Human}, a false positive is considered a GPT-3.5-written comment classified as a human. A false negative is a human-written comment which got classified as GPT-3.5. The analysis is done using SHAP~\cite{lundberg2017unified}. SHAP is a game theoretic approach to explain the output of any machine learning model.\footnote{\url{https://shap.readthedocs.io/en/latest/}} 
While any instance can be analyzed, in this section we consider two misclassified instances, differing in setting and type of misclassification.

\begin{figure}[t]
\begin{center}
\includegraphics[width=0.8\textwidth]{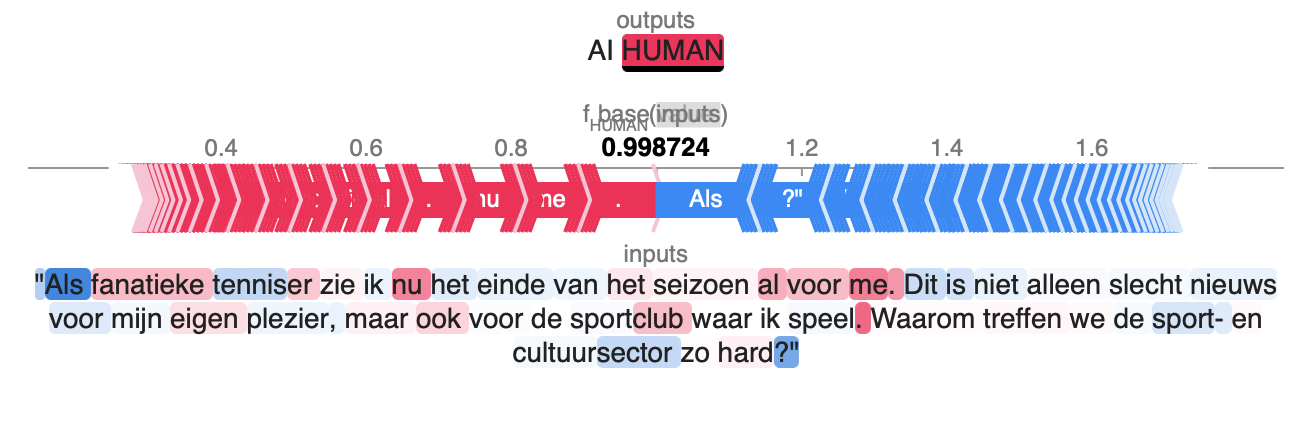}
\end{center}
\caption{False positive: a GPT-generated comment (ZS-1) that was classified as human. ``\textit{As an avid tennis player, I can already see the end of the season ahead of me. This is bad news not only for my own enjoyment, but also for the sports club where I play. Why are we hitting the sports and cultural sector so hard?}''}\label{fig:shapzs1}
\end{figure}

\textbf{False positive}
The first example is a comment on article 1 generated by zero-shot GPT-3.5 as persona 1. In Figure~\ref{fig:shapzs1}, the instance is visualized with SHAP. The model predicted that this particular instance was human-written, but in fact, was generated by GPT-3.5. Especially the first two sentences immediately stand out. The content seems rather personal and sentimental, which is most likely to cause the incorrect classification. This is the perfect example of the model utilizing the given persona, which was the initial intention of providing one. In the original Dutch output it also stood out that GPT-3.5 made a spelling mistake by generating \verb|tenniser| instead of \verb|tennisser|.

As we have selected the assigned human class at the top, the contribution of each token to the human class is shown. The individual contribution of a token is determined by calculating the difference between the total classification and the classification with a single token masked. Tokens in red suggest a positive contribution to the selected class, while tokens in blue suggest a negative contribution. It is evident that tokens in the first two sentences such as \verb|fanatieke|, \verb|het seizoen al voor me.| and \verb|eigen| positively contribute to the classification. The third sentence, which has a rather formal tone, barely contributes positively to the classification. This is in line with our earlier observation. Our model does not associate tokens such as \verb|als|, \verb|tennis|, \verb|sport|, \verb|sector| and \verb|?| with a human-written comment, suggesting these are more GPT-3.5 like.

\begin{figure}[t]
\begin{center}
\includegraphics[width=0.8\textwidth]{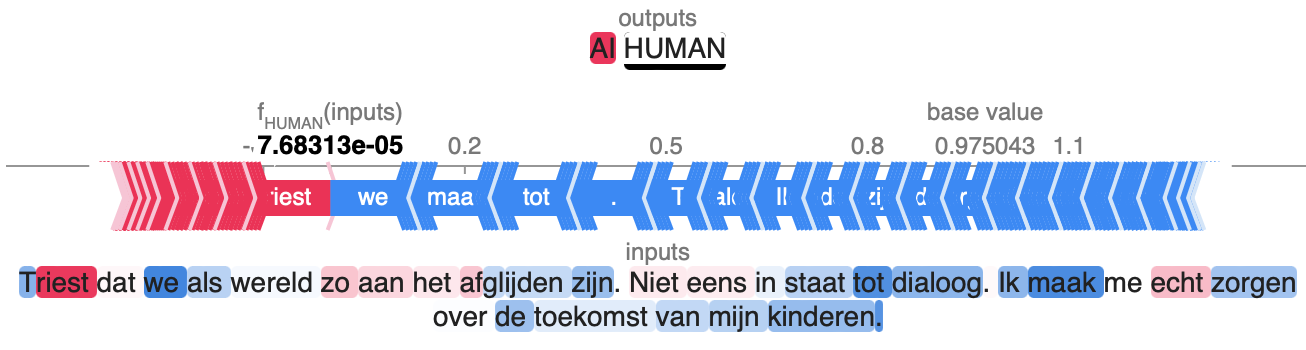}
\end{center}
\caption{False negative: a human comment on article 7 that was classified as GPT-3.5 (CL-2) ``\textit{Sad that as a world, we are slipping away like this. Not even capable of dialogue. I am genuinely worried about the future of my children.}''}\label{fig:shapcl2}
\end{figure}

\textbf{False negative}
A false negative in our case is a human-written comment on article 7 classified as GPT-3.5. In Figure~\ref{fig:shapcl2}, the SHAP values for this instance are shown. 
It is noticeable that the comment has a rather formal structure. Rather than a strong opinion accompanied by personal motivation, the comment presents a rather impersonal perspective through factual statements. The last sentence exhibits the same personal characteristics of P1, which the model potentially associates with GPT-3.5-like opinions. The human class is again selected, meaning that in this scenario all blue tokens contribute positively and all red tokens negatively to the AI class. The first thing that stands out is that almost every token contributes positively to the AI class. It is remarkable that the first word is split into two tokens, which individually do not have a meaning. The token \verb|riest| has the strongest negative contribution of the entire comment. The last sentence has GPT-3.5-like characteristics: Apart from the token \verb|echt|, it strongly contributes to the AI classification.

\section{Discussion}

The results of the fine-tuned BERT models were all very high. This tells us the capabilities of GPT-3.5 are quite limited. Our findings do not suggest that a specific prompting technique (zero-shot, few-shot, or context) results in more human-like outputs. This suggests that capturing the complexity of human-natured comments is still a challenging task for GPT-3.5. The F1 scores are not different enough to make statements about which setting can best reproduce human opinions on news articles. 
We observe that regardless of the specific prompt, GPT-3.5 generally outputs comments which had a rather factual and boring tone of style. We tried to counter this by providing a persona, but this had little to no effect. This was probably because the generated personas were narrowly described and therefore the model had a limited idea about its beliefs and motivations, especially on multiple topics. 
In terms of future work, it will be interesting to experiment with larger amounts of personas, and evaluate the impact compared to not using personas at all.

As we observed earlier that GPT-3.5 generated comments seemed rather formal and often had the same structure, we analyzed the lexical diversity by calculating the CTTR values. We found that human opinions consistently exhibited higher CTTR values, meaning that the ratio of unique words to total words was greater than that of machine-generated comments. 
We saw that few-shot learning resulted in the highest lexical diversity, followed by zero-shot completions. It makes sense that the few-shot setting has the highest value since it directly learns from real-world instances and therefore copies such words more easily. Another finding is that providing the introduction of the article in the prompt reduces the variety of words the model used. However, the difference in diversity among the output of the prompts was much smaller than the difference between human-written or machine-generated comments.


We encountered several limitations during our research. As mentioned in Section \ref{sec:limit1} the API, in fact, had some flaws. On some days, a single completion would take much longer than usual or even not even be generated due to overloaded servers. 
Another issue we had was the token per request limit, not allowing a prompt to run successfully at once. As mentioned earlier, we managed to counter this by catching possible errors in our script. At first, this does not seem like a major issue. However, we found differences in the outputs of the prompt. While being prompted the same, the output differed between requests due to the probabilistic behaviour of the generative model. Within one request batch, the style was consistent though, e.g. 
the adding of quotation marks or starting every comment with the same words. 

\section{Conclusion}

The goal of this research was to investigate to what extent GPT-3.5 can generate human-like comments on Dutch news articles and how to best generate these. We answered this question by experimenting with multiple prompting techniques, after which we could analyze the different outputs. In particular, the zero-shot, few-shot and context prompts were utilized, corresponding with a generated persona. We fine-tuned the pre-trained RobBERT-v2 model to classify whether unlabeled comments were human-written or generated by GPT-3.5.

While in previous research zero-shot and few-shot learning had shown remarkable performances, it does not so in our case. We found that the BERT models we fine-tuned were able to achieve high classification scores. 
We can conclude that GPT-3.5 is still limited in generating human-like comments on Dutch news articles, regardless of which prompting setting. It suggests that capturing the complexity of human-nature comments, even with real examples, is still a challenging task.

One of the findings from our analysis is that human-written comments generally have a much higher lexical diversity than GPT-3.5-generated comments. Although the differences are small, few-shot prompts averaged the highest lexical diversity but still lower than human comments. The manual analysis of individual misclassifications led to additional insights that GPT-3.5 very often tends to generate comments in a rather formal and factual style and less opinionated than humans.

During our research, OpenAI publicly announced GPT-4.\footnote{\url{https://openai.com/gpt-4}} This may be an advantage in future studies on opinionated text generation. Instead of the current limit of 4,096 tokens, GPT-4 is capable of handling 25,000 tokens per request. This is a major improvement and can be used to run prompts at once, instead of running them in batches. Apart from GPT-4, we encourage follow-up research with open-source LLMs such as BLOOM. Secondly, it can be used to provide more context such as the entire article in the prompt, which would possibly lead to more in-depth comments. At last, a lot more human-written examples can be provided in the prompt, which may positively influence the human-likeness of the output. Other potential further research direction is the investigation of to what extent the pre-training date of GPT-3.5 influences its performance, experimenting with subjects from different time periods before and after this date.

\bibliographystyle{plain}
\bibliography{bibliography.bib} 
\end{document}